\documentclass[journal,12pt,onecolumn,draftclsnofoot]{IEEEtran}
\usepackage{xcolor,soul,framed} 
\usepackage{setspace}

\colorlet{shadecolor}{yellow}
\usepackage[pdftex]{graphicx}
\graphicspath{{../pdf/}{../jpeg/}}
\DeclareGraphicsExtensions{.pdf,.jpeg,.png}

\usepackage[cmex10]{amsmath}
\usepackage{array}
\usepackage{mdwmath}
\usepackage{mdwtab}
\usepackage{eqparbox}
\usepackage{url}
\usepackage{algorithm}
\usepackage{algpseudocode}
\usepackage[percent]{overpic}
\usepackage{amsmath}
\usepackage{amsfonts}
\usepackage{authblk}
\usepackage{hyperref}

\begin{document}

\title{3D Neuron Morphology Analysis}
\author{Jiaxiang Jiang$^{\star}$, Michael Goebel$^{\star}$, Cezar Borba$^{\dagger}$, William Smith$^{\dagger}$  and B.S. Manjunath$^{\star}$} 
\affil{$^{\star}$ Department of Electrical and Computer Engineering, University of California, Santa Barbara \\$^{\dagger}$ Department of Molecular, Cell and Developmental Biology, University of California, Santa Barbara
}

\maketitle

\begin{abstract}
We consider the problem of finding an accurate representation of neuron shapes, extracting sub-cellular features, and classifying neurons based on neuron shapes.
In neuroscience research, the skeleton representation is often used as a compact and abstract representation of neuron shapes.
However, existing methods are limited to getting and analyzing \textquotedblleft curve\textquotedblright skeletons which can only be applied for tubular shapes. 
This paper presents a 3D neuron morphology analysis method for more general and complex neuron shapes.
First, we introduce the concept of skeleton mesh to represent general neuron shapes and propose a novel method for computing mesh representations from 3D surface point clouds.
A skeleton graph is then obtained from skeleton mesh and is used to extract sub-cellular features.
Finally, an unsupervised learning method is used to embed the skeleton graph for neuron classification.
Extensive experiment results are provided and demonstrate the robustness of our method to analyze neuron morphology.

\end{abstract}

\begin{IEEEkeywords}
3D Neuron Morphology, Skeleton Mesh, Graph, Sub-cellular features, Embedding, Classification
\end{IEEEkeywords}

%


\section{Introduction}
\begin{figure*}[ht]
    \centering
    \includegraphics[width=0.5\textwidth ]{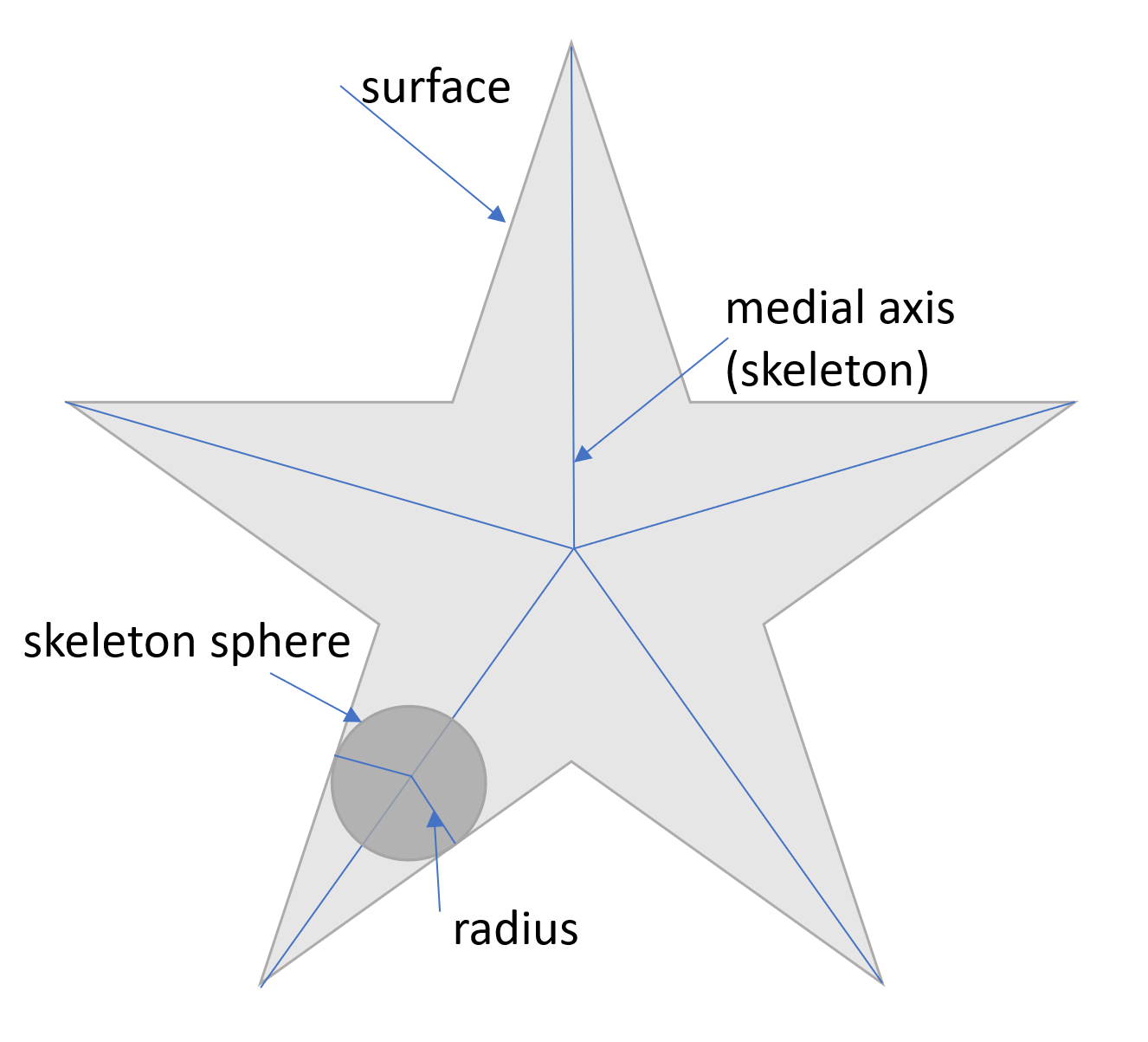}
    \caption{Illustration of MAT by using a 2D shape example.}
    \label{fig:MAT_definition}
\end{figure*}

The importance of neuronal morphology has been recognized from the early days of neuroscience \cite{neuronmorphologyimpoartance}.
There are three obstacles in automatic neuron morphology analysis. 
First, we need to have a good shape representation of each neuron.
Skeleton representations are widely used in neuroscience \cite{neuron_tracing,skeleton_1_Navis,skeleton_2_neuromopho,skeleton_3} as they provide a compact and abstract shape representation.
Mathematically, skeletonization or medial axis transform (MAT) has a rigorous definition for arbitrary shapes.
The skeleton of a shape is defined as a collection of interior points that have at least two closest points on the surface of the shape. 
We refer to those interior points as skeleton points and each skeleton point is associated with a radius.
Figure \ref{fig:MAT_definition} shows an example of MAT.  
However, in reality, it is not an easy task to get skeleton representation directly from images.
Most automatic or manual segmentation methods output a cloud of surface points.
Thus, we need to compute the 3D neuron skeleton from 3D surface point clouds.
The skeleton representation further enables computing sub-cellular features such as length and number of branches of neurons as well as classification of neurons.

The main contribution of this paper is a robust and efficient method for computing a skeleton representation from a set of 3D surface points. This 3D skeleton representation can be used for a quantitative analysis of neuronal cell structures, including sub-cellular feature calculations and for neuron type classification based on 3D shapes.

\begin{figure*}[ht]
    \centering
    \includegraphics[width=0.5\textwidth ]{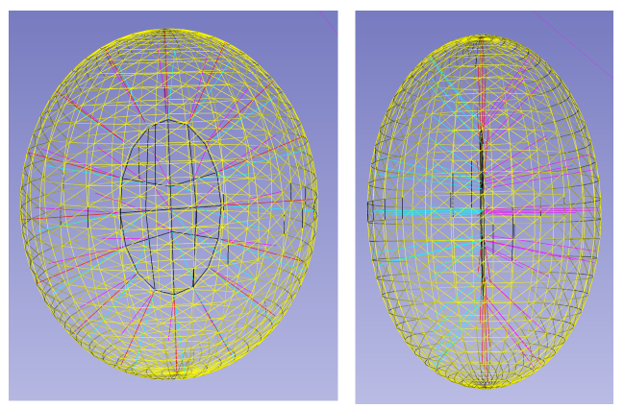}
    \caption{Visualization of a 3D ellipsoid shape and its surface skeleton from two points of view. Yellow triangle mesh represents object surface. Black contour represents the outline of the skeleton surface. Magenta and Cyan line segments represent two closest surface points from the skeleton point. Two colors are used to differentiate different directions.}
    \label{fig:MAT_surface}
\end{figure*}

There is an extensive literature on neuron skeleton extraction \cite{skeleton_1_Navis,skeleton_method2,unet_skeleton}.
In \cite{skeleton_1_Navis}, the skeleton representation is computed from a 3D mesh by using a traditional morphological thinning algorithm \cite{thinning_algorithm}. 
This method has two main drawbacks.
First, the thinning algorithm is sensitive to noise of 3D mesh.
Second, in reality, we usually get discrete 3D surface points of neurons from the segmentation step, and constructing 3D mesh from those discrete 3D surface points will introduce additional noise.
To make the skeleton extraction model more robust, \cite{unet_skeleton} proposes to use deep learning network to learn skeleton representation.
The main idea of the paper is to use the deep learning network to predict skeletons from features of multiple spatial scale layers.
This model still takes a continuous surface as input, as opposed to discrete surface points.
Further, this is a supervised method and it requires training samples. 
In \cite{skeleton_method2}, they propose extracting skeleton representations directly from discrete surface points by using a 3D discrete distance transform.
However, this only works well for curve skeletons and only tubular structures have curve skeletons.
General 3D shapes will result in surface skeletons as shown in Fig.\ref{fig:MAT_surface}.
In practice, skeleton mesh is used to represent the surface skeleton.
There is no existing method to extract skeleton mesh from surface point clouds for neuron morphology analysis. 

There are also methods to analyze neuron shapes using the skeleton representation. For skeleton classification task, \cite{skeleton_1_Navis} proposes to compare similarity of skeletons by using local skeleton features.
It breaks a neuron skeleton into short segments and characterize segments by location and direction of segments.
However, this method only works well for skeleton curves but not surface skeletons.
In \cite{s_rep_object}, they only consider features from skeleton points for the classification and it does not fully utilize the skeleton mesh information.

To analyze general neuron shapes, this paper presents a robust 3D neuron morphology analysis framework based on the surface skeleton representation of neurons. 
In \cite{lin2021point2skeleton}, the authors propose an unsupervised deep learning skeleton mesh extraction method.
However, this method does not work well when neurons have concave shapes.
Our skeleton mesh extraction method is built upon \cite{lin2021point2skeleton}, and by using estimated surface norm of point clouds as part of the optimization function, we address this drawback.
Next the skeleton mesh is converted to an undirected graph called skeleton graph.
Inspired by \cite{sun2019infograph}, we embed the skeleton graph by maximizing mutual information, and then classify neurons based on the embedding of each skeleton.
To compute cellular/sub-cellular features of neurons from the skeleton representation, we also utilize the skeleton graph.
A simple but effective recursive algorithm is proposed to get number of branches and length of neurons.

We apply our neuron morphology analysis method to classify \textit{Ciona} neurons. The \textit{Ciona} sea squirt is one of the widely studied tunicates in neuroscience \cite{kerrianne2016cns}. 
Its brain is closely related to vertebrates with a much simpler neuronal structure.
In a single \textit{Ciona} larva, it has only about 187 neurons with about 6600 synapses \cite{kerrianne2016cns}. 
Studying the \textit{Ciona} brain in depth can reveal the general principles behind the mechanism of how vertebrate brains work \cite{scheffer2020connectome}.
We also present our results on the NeuroMorpho\cite{ascoli2007neuromorpho} dataset.   
In summary, the main contribution of our paper include
\begin{itemize}
    \item A robust and efficient skeleton mesh extraction method with novel cost function by using properties of MAT. To the best of our knowledge, this is the first one to use skeleton mesh instead of the curve skeleton to analyze neuronal shapes
    \item A 3D \textit{Ciona} neuron dataset that can be used for neuron morphology analysis.
\end{itemize}

\section{Method}
\begin{figure*}[ht]
    \centering
    \includegraphics[width=1.0\textwidth ]{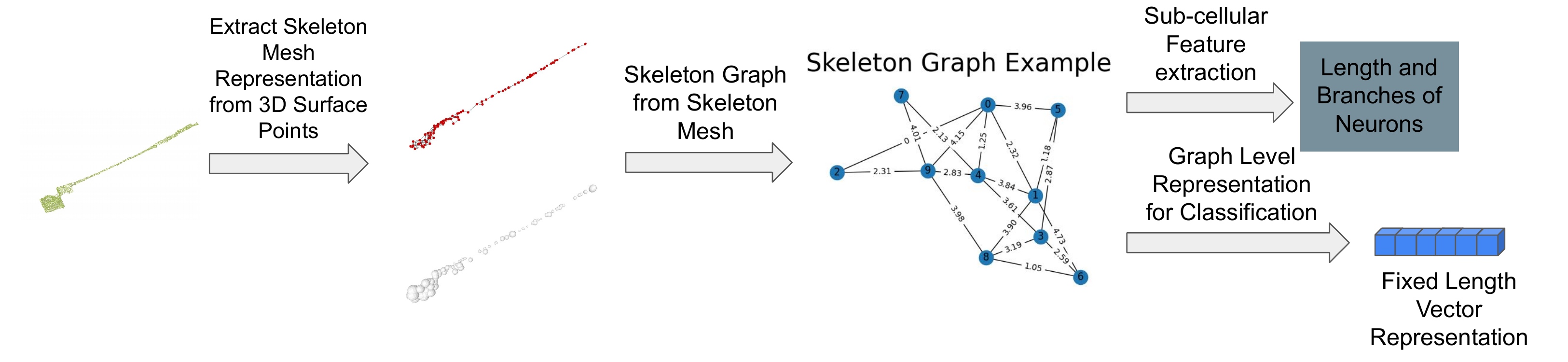}
    \caption{Overview of our proposed neuron morphology analysis pipeline. Given a surface point cloud as input, we extract the skeleton mesh. The skeleton mesh includes skeleton points with their radii as well as the connection of skeleton points. Then we construct the skeleton graph. Each node in a skeleton graph represents a skeleton point, and edge in the graph represents the connection between skeleton points.  Next, we propose a graph analysis method to get length and number of branches of neurons based on the skeleton graph. We also use the skeleton graph for classification task by embedding it into a fixed length vector.}
    \label{fig:workflow}
\end{figure*}

Fig.\ref{fig:workflow} illustrates our overall neuron morphology analysis method. 
Given a set of surface point clouds as input, we introduce an unsupervised deep learning method to get the skeleton mesh representation of each neuron.
This is achieved by using the properties of the traditional medial axis transform (MAT).
The skeleton mesh representation includes skeleton points with radii as well as the connection of those skeleton points as shown in Fig.\ref{fig:workflow}.
Second, the skeleton representation of each neuron is transformed into a skeleton graph.
Each node in the skeleton graph represents a skeleton point.
If there is an edge between two nodes, it means those two skeleton points are connected.
The weight of the edge represents distance between the two skeleton points.
Radii as well as the location of each skeleton point are attributes of each node.
Next, length and number of branches of neurons are computed from the skeleton graph.
To compare different shapes of neurons, a graph level representation learning method is used to embed the skeleton graph.
The representation learning method is an unsupervised method that maximizes mutual information of the skeleton graph.

\subsection{Skeleton Mesh from 3D Surface Point Cloud}
\begin{figure*}[ht]
    \centering
    \includegraphics[width=1.0\textwidth ]{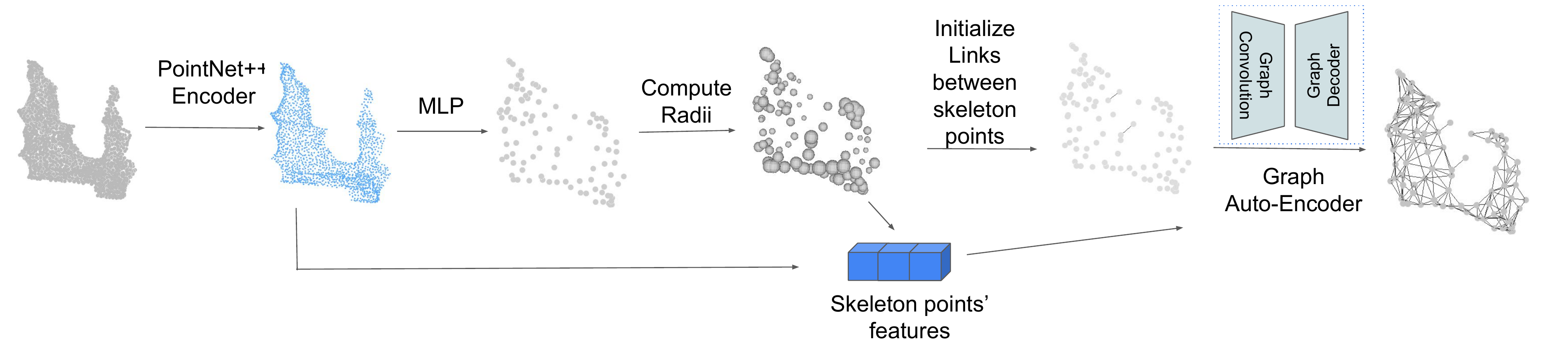}
    \caption{Overview of neuron skeleton representation method. Given 3D surface point cloud as input, PointNet++\cite{qi2017pointnet++} is used to extract features of the input point cloud. Then a geometric transformation learned by MLP will predict the skeleton points location with their radii. After skeleton points prediction, two simple priors are used to initially connect some skeleton points, and a graph auto-encoder is used to predict all links that connect skeleton points.} 
    \label{fig:skeleton_extraction}
\end{figure*}
Our unsupervised 3D neuron skeleton extraction method is built upon the method in \cite{lin2021point2skeleton} as illustrated in Fig.\ref{fig:skeleton_extraction}.
Given a 3D surface point cloud as input, PointNet++ \cite{qi2017pointnet++} is used as the encoder to obtain the sampled surface points with features.
Next, a multi-layer-perceptron (MLP) is used to learn the geometric transformation to predict the skeleton points with their radii using linear combination of the MLP input points.
Compared to \cite{lin2021point2skeleton}, we propose to use properties of skeleton points as the prior knowledge to make the geometric transformation learning process more robust to general shapes.
After getting skeleton points with radii, a graph auto-encoder is used to predict links between skeleton points.

\subsubsection{skeleton points prediction}

Mathematically, given a set of 3D surface points $\textbf{P}\in \mathbb{R}^{M\times 3}$ where $M$ is a number of points, we want to predict $N$ skeleton spheres $\textbf{s}_i=<\textbf{c}_i,r_i>$ where $\textbf{c}_i$ is the center of each sphere and $r_i$ is the radius of each sphere.

As illustrated in Fig.\ref{fig:skeleton_extraction}, we first use PointNet++\cite{qi2017pointnet++} as the encoder to obtain a set of sampled input points $\textbf{P}'\in \mathbb{R}^{M'\times 3}$ and their associated contextual features $\textbf{F}\in\mathbb{R}^{M'\times D}$. $M'(M'<M)$ is the number of feature points after PointNet++ and $D$ is the feature dimension of each feature point. 
PointNet++ groups points and extract point features hierarchically. It contains a number of set abstraction levels. 
For each set abstraction level, there are three layers: Sampling layer, Grouping layer, and PointNet layer. 
For the first set abstraction level, the input is $\textbf{P}$, a set of $M$ number of 3D surface points. 
Next, Sampling layer is applied. 
The iterative farthest point sampling (FPS) \cite{qi2017pointnet++} is used to get $M'$ sampled points. 
In the grouping layer, those $M'$ sampled points are used as the centroid points. 
Then for each centroid, all $M$ points within a radius are viewed as neighbor points and are grouped into that local region. 
Therefore, each centroid has $K$ neighbors and $K$ can vary for different groups. 
After Sampling and Grouping layers, PointNet\cite{qi2017pointnet} layer is used to extract features for each local region.
The sampling layer, grouping layer, and PointNet layer consists one set abstraction level, and we stack such abstraction levels to form a hierarchical architecture to get features at different spatial scales.
Next, multi-scale grouping is applied to concatenate the features from different spatial scales.

A Multi-Layer Perceptron (MLP) is used to estimate the geometrical transformation to get a set of skeleton spheres' center points $\textbf{C}$, $\{ \textbf{c}_1,\textbf{c}_2,...,\textbf{c}_N \}$. 
The geometric transformation we use is a convex combination of input points $\textbf{P}'$.
MLP with softmax function is used to estimate the weight $\textbf{W}\in \mathbb{R}^{M'\times N}$ of the convex combination in eq. \ref{eq:convex_skeleton}.
 \begin{equation}\label{eq:convex_skeleton}
    \textbf{C}= \textbf{W}^T\textbf{P}' \qquad \text{subject to} \qquad \forall j\in \{ 1,...,N \} \quad \sum_{i=1}^{M'}W_{i,j}=1
 \end{equation}

As shown in \cite{lin2021point2skeleton}, the same weight matrix \textbf{W} can be used to compute $r(\textbf{c})\in \textbf{R}$  using eq.\ref{eq:radii_compute}

 \begin{equation}\label{eq:radii_compute}
    \textbf{R}= \textbf{W}^T\textbf{D}
 \end{equation}

\noindent where $D\in\mathbb{R}^{M'\times 1}$ is a vector of  $d(\textbf{p}',\textbf{C})$. $d(\textbf{p}',\textbf{C})$ is the closest distance of one surface point $\textbf{p}'$ to all skeleton points $\textbf{C}$ and is defined as $d(\textbf{p}',\textbf{C})=\min_{\textbf{c}\in\textbf{C}}||\textbf{p}'-\textbf{c}||_2$

A set of loss functions are defined in \cite{lin2021point2skeleton} to train the MLP.
The loss function includes sampling loss $L_s$, point-to-sphere loss $L_p$, and radius regularizer loss $L_r$. 
The first two losses are based on the recoverability of skeleton representation. The last loss term is to encourage larger radii to avoid instability induced by surface noise. 

For the sampling loss $L_s$, we sample points on the surface of each skeleton sphere and measure the Chamfer Distance (CD) between  the set of sampled sphere points $\textbf{T}$ and the set of sampled surface points from PointNet++ $\textbf{P}'$ as in eq. \ref{eq:sampling_loss}:
\begin{equation}\label{eq:sampling_loss}
    L_s= \sum_{\textbf{p}'\in\textbf{P}'}\min_{\textbf{t}\in\textbf{T}}||\textbf{p}'-\textbf{t}||_2 + \sum_{\textbf{t}\in\textbf{T}}\min_{\textbf{p}'\in\textbf{P}'}||\textbf{t}-\textbf{p}'||_2
 \end{equation}

Point-to-sphere loss $L_p$ measures the reconstruction error by explicitly optimizing the coordinate of skeleton points and their radii:
\begin{equation}\label{eq:point_to_spere}
    L_p= \sum_{\textbf{p}'\in\textbf{P}'}(\min_{\textbf{c}\in\textbf{C}}||\textbf{p}'-\textbf{c}||_2-r(\textbf{c}_{p'}^{min}) )+ \sum_{\textbf{c}\in\textbf{C}}(\min_{\textbf{p}'\in\textbf{P}'}||\textbf{c}-\textbf{p}'||_2-r(\textbf{c}) )
 \end{equation}
where $\textbf{C}$ is a set of predicted skeleton points, $r(c)$ is a radius of skeleton point $\textbf{c}$, and $\textbf{c}_{p'}^{min}$ is the cloest skeleton points to a point $\textbf{p}'$.

Radius regularizer loss $L_r$ is defined in eq.\ref{eq:radius_regularizer} where $r(\textbf{c})$ is a radius of the skeleton point $\textbf{c}$. By minimizing this loss, it encourages large radii of skeleton points to make the skeleton points prediction more stable.
\begin{equation}\label{eq:radius_regularizer}
    L_r= -\sum_{\textbf{c} \in \textbf{C}} r(\textbf{c})
 \end{equation}

\begin{figure*}[ht!]
    \centering
    \includegraphics[scale=0.8]{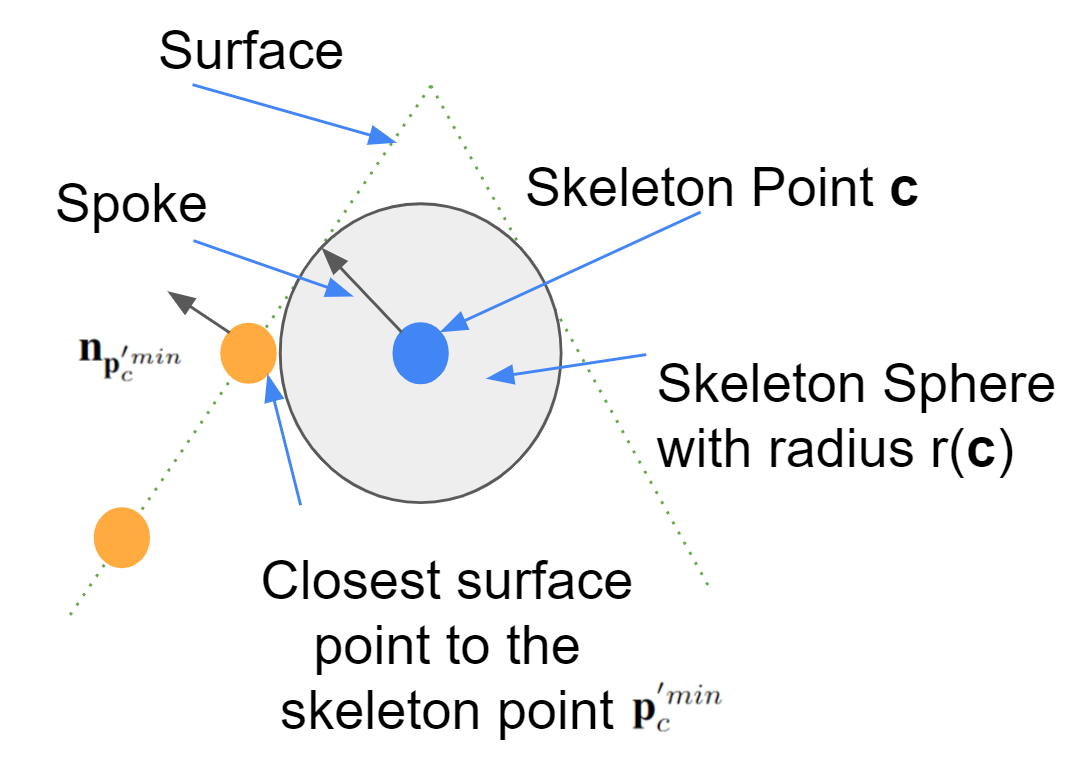}
    \caption{Spoke is a vector connecting a skeleton point and that skeleton point's one of two closest surface points. The vector points from the skeleton point to the surface point. Green dashed lines represent the implicit surface of an object, blue dot is one skeleton point, orange dots represent surface points, and the arrow represents a spoke. Spoke is perpendicular to the object surface at the surface point.}
    \label{fig:spoke and norm}
\end{figure*}

However, based on above three losses, predicted skeleton points can be outside of a shape if the shape is concave. Therefore, we introduce the skeleton-to-surface norm loss $L_n$ to encourage the skeleton points to be inside the shape. $L_n$ is a term to measure the reconstruction error by utilizing the property of spoke direction in MAT. 
Fig.\ref{fig:spoke and norm} illustrates a spoke of a skeleton point in MAT.
The length of a spoke of the skeleton point $\textbf{c}$ is $r(\textbf{c})$.
This is also one of our main contribution compared to \cite{lin2021point2skeleton} for skeleton points prediction.
In theory, the direction of the spoke should be perpendicular to the object surface at the surface point \cite{s_rep_object}.
Also the spoke direction should be pointing outside of a shape.
To capture this property, we define $L_n$:

\begin{equation}\label{eq:norm_loss}
    L_n= \sum_{\textbf{c}\in\textbf{C}}(1- \textbf{n}_{\textbf{p}_c^{'min}}\cdot\frac{\textbf{p}_c^{'min}-\textbf{c}}{||\textbf{p}_c^{'min}-\textbf{c}||_2})+\sum_{\textbf{p}'\in\textbf{P}'}(1- \textbf{n}_{p'}\cdot\frac{\textbf{p}'-\textbf{c}_{p'}^{min}}{||\textbf{p}'-\textbf{c}_{p'}^{min}||_2})
 \end{equation}

\noindent $\textbf{p}_c^{'min}$ is the closest surface point to the skeleton point $c$ and $\textbf{n}_{\textbf{p}_c^{'min}}$ is the estimated surface norm of the surface point. The ``$\cdot$'' denotes the dot product between two vectors. To estimate the norm of each point in the 3D surface points, the adjacent points are found first and then principal axis of the adjacent points using covariance analysis are calculated. More details of the norm estimation of each surface point are described in \cite{zhou2018open3d}. $L_n$ encourages the skeleton points to be located within a shape even the shape is concave. 

\subsubsection{Links prediction}

After predicting skeleton points, our target is to predict links to connect skeleton points to form the skeleton mesh. 
In theory, for any pair of skeleton points, if all points that are on the line connecting the two skeleton points are also on the skeleton surface, there should be links between those two points. We adapt the graph auto encoder (GAE) as used in \cite{lin2021point2skeleton} to predict links between skeleton points. GAE takes input the initialized adjacency matrix $A_{ini}$ of the skeleton mesh graph $G_{mesh}$ and the skeleton points' features. The skeleton points' features is concatenation of $\textbf{C}$, $\textbf{R}$, and $\textbf{W}^T\textbf{F}$.  $\textbf{C}$ are coordinates of skeleton points, $\textbf{R}$ are radii of skeleton points, $\textbf{W}$ is the learned weights from the MLP, and $\textbf{F}$ is the contextual features of the surface points from PointNet++. GAE outputs the estimated adjacency matrix $\hat{A}$ of $G_{mesh}$. The loss function is a Masked Balanced Cross-Entropy (MBCE) loss as proposed in \cite{loss_GAE}.

\subsection{Sub-cellular feature extraction from the skeleton graph}

Skeleton model can be represented as the skeleton graph $G(V,E)$ where $V$ represents all skeleton points and $E$ represents connection between skeleton points. 
The weight of the edge represents distance between skeleton points.  

\subsubsection*{Neuron length computation from skeleton graph}

We formulate the neuron length computation problem as finding the longest simple path in the skeleton graph $G$.
A simple path in the graph is a path that does not have repeat nodes.
In $G$, the length of the path is the sum of all edges' weights along the path. 
Note that our skeleton graph can have loops. 
To solve this problem, for each node, we find the longest simple path from that node and denote as $path(i)$ for node $i$.
To avoid getting stuck during the loop, we mark any node when we visits as shown in the algorithm below.
Next, we find $i^*$ that maximizes $path$. 
We use the following recurrent algorithm to find the longest simple path from one node in the skeleton graph.
\begin{algorithm}
   \caption{Find the longest simple path from node i}
    \begin{algorithmic}
    \State D[j] represents the longest path from j to i. Initially, D[j]=0 for all j 
      \Function{LongestPath}{i, currLength}
        \If{Node i is visited}
        \State return
        \EndIf
        \State change i status to visited
        \If{D[i] \textless currLength}
        \State D[i]=currLength
        \EndIf
        \For {all nodes j that is connected to i} 
        \State LongestPath(j, currLength+edgeweight[i,j])
        \EndFor
        \State change i status to not visited
       \EndFunction
\end{algorithmic}
\end{algorithm}

\subsubsection*{Neuron branch calculation}

After finding the longest simple path, we are able to identify a set of nodes on that path.
Those nodes are possible branching nodes.
We name a set containing all possible branching nodes as $\textbf{B}$
For each node $i\in B$, we find the longest simple path from that node $i$ which does not contain any other nodes in $B$.
Therefore, the branch is identified as the longest simple path.

\subsection{Skeleton Model Comparison}

\begin{figure*}[ht!]
    \centering
    \includegraphics[width=1.0\textwidth]{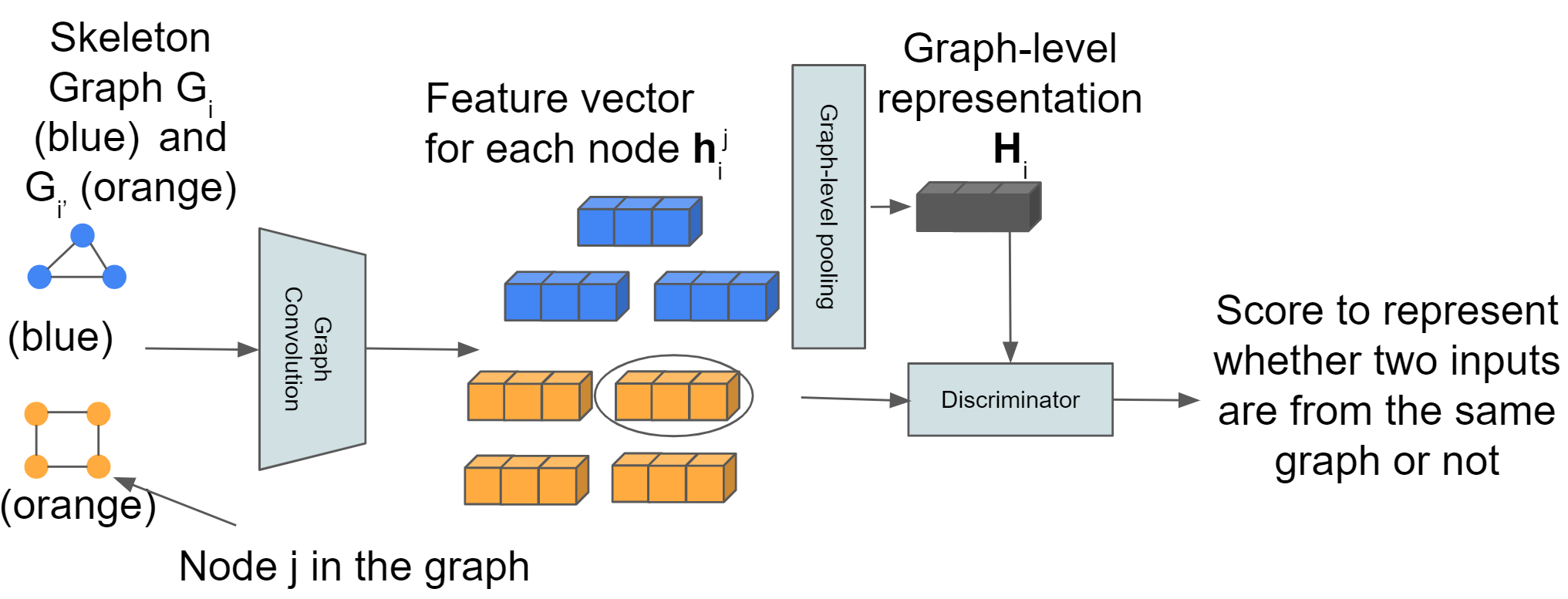}
    \caption{We use two example skeleton graphs (blue and orange) to demonstrate how we embed the skeleton graph. Each node of a skeleton graph is encoded into a feature vector by using graph convolution layers. A fixed length graph level feature vector (global representation) is obtained by graph-level pooling operation of each node feature vector. The discriminator takes inputs both global representation and patch representation to decide whether they are from the same skeleton graph. In this toy example, there will be 14 global-patch pairs.}
    \label{fig:infograph}
\end{figure*}

We cluster neuron morphology by comparing different skeleton graphs. Specifically, we embed the skeleton graphs and then cluster neurons based on their embeddings. We embed the skeleton graph based on InfoGraph \cite{sun2019infograph} as illustrated in Fig. \ref{fig:infograph}. The embedding process is in an unsupervised manner. 

First, graph convolutional layers are used to generate node features which we refer to as patch representation $\textbf{h}_i^j$ ($i$ is the skeleton graph index and $j$ is the node index of the skeleton graph $i$).
Then graph-level pooling is used on all patch representations to get the graph level representation (global representation) $\textbf{H}_i$.
The mutual information (MI) estimator on global-patch pairs over the given graph dataset $\textbf{G}$ is defined as:
\begin{equation}\label{eq:mutual_info}
    MI= \sum_{i\in K}\frac{1}{K}\sum_{j \in G_i}I(\textbf{h}_i^j,\textbf{H}_i)
 \end{equation}
where $K$ is the total number of graphs in the dataset and $G_i\in \textbf{G}$.

MI is the mutual information estimator modeled by the discriminator $T$. The Jensen-Shannon MI estimator proposed in \cite{sun2019infograph} is
\begin{equation}\label{eq:MI_estimator}
    I(\textbf{h}_i^j,\textbf{H}_i) = \mathbb{E}[-sp(-T(\textbf{h}_i^j,H_i)]-\mathbb{E}[-sp(-T(\textbf{h}_{i'}^j,H_i)]
 \end{equation}
where $\mathbb{E}$ is the expectation (here it is just average operation) and $sp(z)=log(1+e^z)$. $i$ and $i'$ denote two graph samples from the dataset $\textbf{G}$.
The discriminator $T$ estimates global-patch representation pairs by passing two representations to different non-linear transformations and then takes the dot product of the two transformed representations.
Both non-linear transformations consist 3 linear layers with ReLU activation functions.
Therefore, the discriminator will output a score between $[0,\infty)$ to represent whether the input patch (node) is from the input graph.
If the input global/patch pairs are from the same graph, we refer to them as positive samples, otherwise negative samples.
We randomly sample pairs as input to the discriminator.

\section{Dataset}
\subsection{\textit{Ciona} Neuron EM Dataset}
The first dataset (Dataset 1) contains two \textit{Ciona} larva 3D TEM images \cite{kerrianne2016cns}. 
The section thickness for TEM images varies between 35 nm and 100 nm. 
For each section, xy resolution is 3.85$\times$3.85 nm. 
Animal 1 contains 7671 sections and animal 2 contains about 8000 sections. 
In each \textit{Ciona} larva, 187 neurons are annotated. 
Those 187 neurons can be grouped into 31 types.
For animal 2, \textit{Ciona} neuron skeletons are traced using TrackEM2 \cite{cardona2012trakem2}, an ImageJ \cite{collins2007imagej} plugin.
This dataset is summarized in Table \ref{tab:Ciona_dataset}, and we refer to \cite{kerrianne2016cns} for more details.

\begin{table*}[ht]
\scriptsize
    \centering
    \caption{Details of \textit{Ciona} Dataset. It contains two \textit{Ciona} animals, one with surface point cloud annotated and one with skeleton annotated. }
    \begin{tabular}{|c|c|c|c|c|}
    \hline
         Animal& xy resolution (nm)&section thickness (nm)&number of sections& annotations\\
         \hline
         Animal 1 & 3.85$\times$3.85 &35-100 &7671&3D surface point cloud of neurons are provided \\
        \hline
        Animal 2 & 3.85$\times$3.85 &35-100 & 6928 &3D neuron skeletons without skeleton points' radii \\
        \hline
    \end{tabular}
    \label{tab:Ciona_dataset}
\end{table*}

\subsection{\textit{C.elegans} Neuron Dataset from NeuroMorpho}
NeuronMorpho \cite{ascoli2007neuromorpho} is a publicly available dataset that is used for neuron morphology research. 
It has dozens of different animals' neurons.
So far, it is the largest neuron skeletons dataset with associated metadata. 
In this paper, we take a subset of \textit{C.elegans} dataset (Dataset 2) from the whole NeuroMorpho dataset to verify our sub-cellular feature extraction method.
Dataset 2 consists of 299 neuron skeletons (with radii) and it is classified into 10 different types.
Each neuron with detailed metadata information such as number of branches and length of neuron.


\section{Results}

\subsection{Skeleton Model from 3D Surface Point Cloud}
We apply our method on animal 1 neurons from Dataset 1 for the purpose of building a shape model to analyze neuron morphology.
To get the fixed number of 3D surface input points, we use the sampling strategy described in \cite{poisson_sample}.
The main idea of the sampling strategy is to give each point a weight based on its distance to neighbor points. 
Then we sample points based on the weights until we reach the number of desired points. 
Details of defining the neighbor points and computing the weights are described in \cite{poisson_sample}.

\begin{figure*} [ht]
\centering

\begin{overpic}[width=0.2\textwidth]{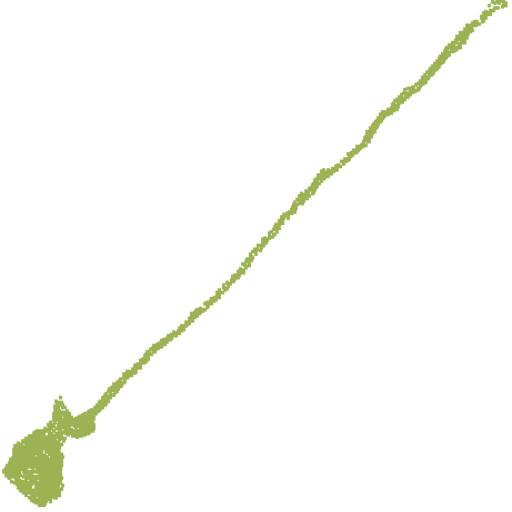}\put(0,80){A}\end{overpic}
\begin{overpic}[width=0.2\textwidth]{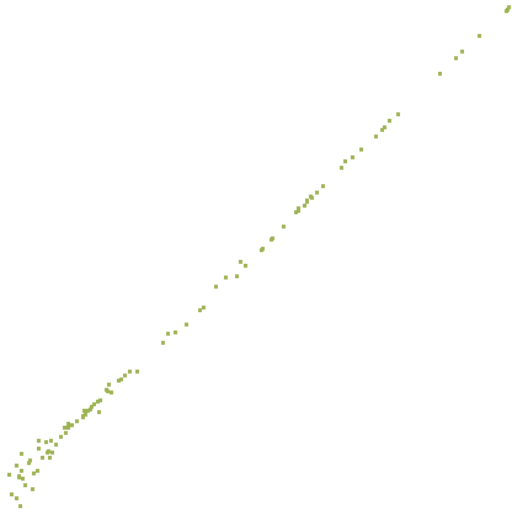}\put(0,80){B}\end{overpic}
\begin{overpic}[width=0.2\textwidth]{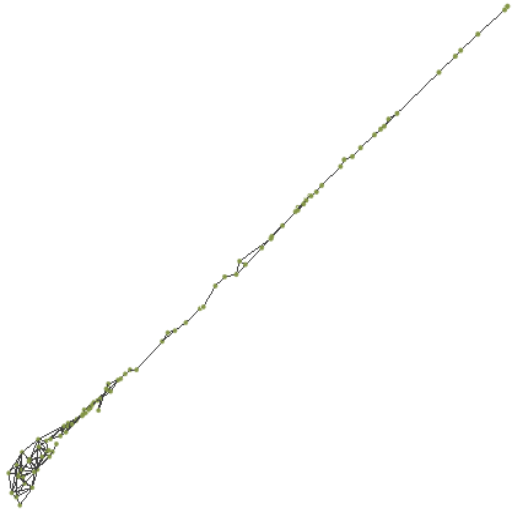}\put(0,80){C}\end{overpic}
\begin{overpic}[width=0.2\textwidth]{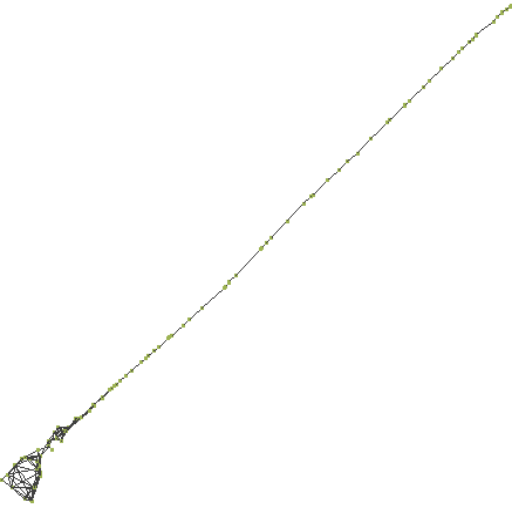}\put(0,80){D}\end{overpic}
\vspace{15pt}

\begin{overpic}[width=0.2\textwidth]{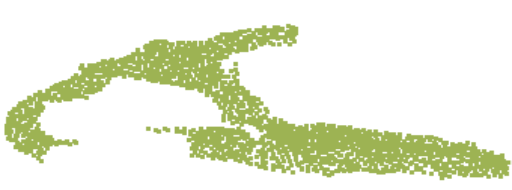}\put(0,30){A}\end{overpic}
\begin{overpic}[width=0.2\textwidth]{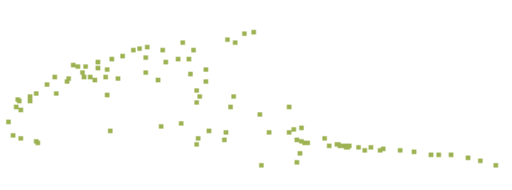}\put(0,30){B}\end{overpic}
\begin{overpic}[width=0.2\textwidth]{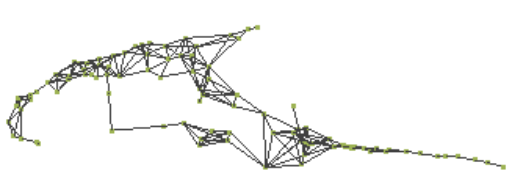}\put(0,30){C}\end{overpic}
\begin{overpic}[width=0.2\textwidth]{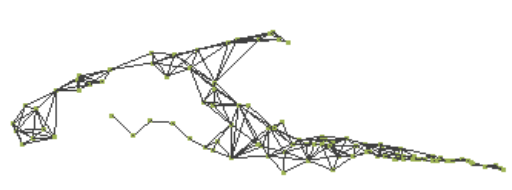}\put(0,30){D}\end{overpic}
\vspace{15pt}

\begin{overpic}[width=0.2\textwidth]{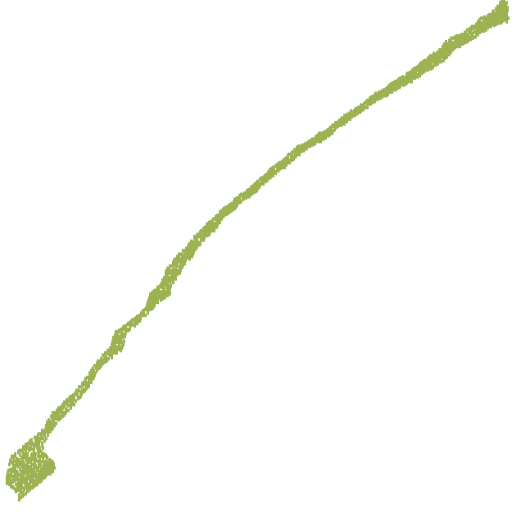}\put(0,80){A}\end{overpic}
\begin{overpic}[width=0.2\textwidth]{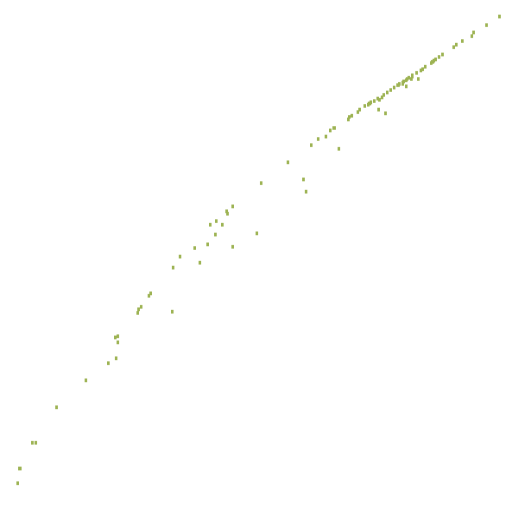}\put(0,80){B}\end{overpic}
\begin{overpic}[width=0.2\textwidth]{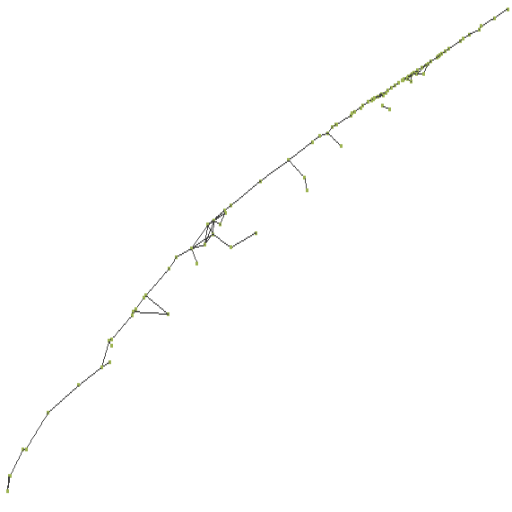}\put(0,80){C}\end{overpic}
\begin{overpic}[width=0.2\textwidth]{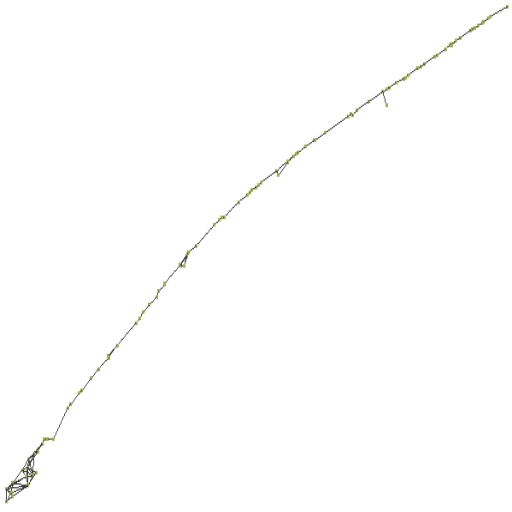}\put(0,80){D}\end{overpic}
\vspace{15pt}

\begin{overpic}[width=0.2\textwidth]{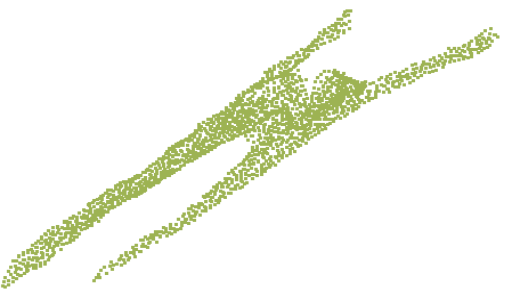}\put(0,40){A}\end{overpic}
\begin{overpic}[width=0.2\textwidth]{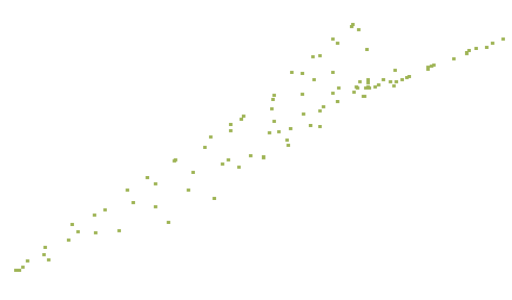}\put(0,40){B}\end{overpic}
\begin{overpic}[width=0.2\textwidth]{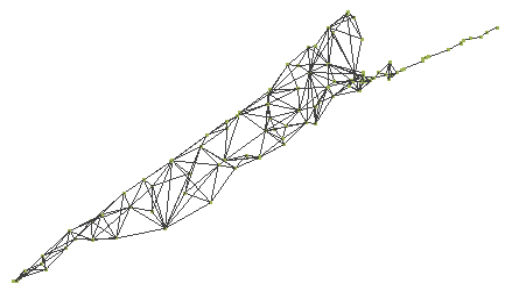}\put(0,40){C}\end{overpic}
\begin{overpic}[width=0.2\textwidth]{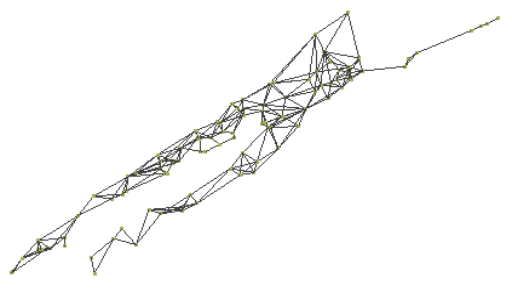}\put(0,40){D}\end{overpic}
\vspace{15pt}

\begin{overpic}[width=0.2\textwidth]{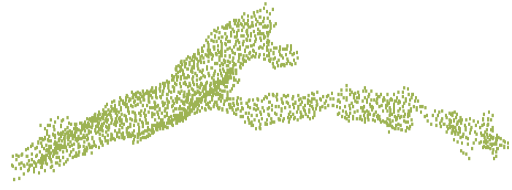}\put(0,30){A}\end{overpic}
\begin{overpic}[width=0.2\textwidth]{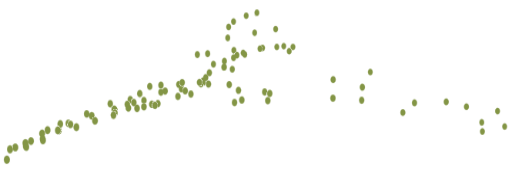}\put(0,30){B}\end{overpic}
\begin{overpic}[width=0.2\textwidth]{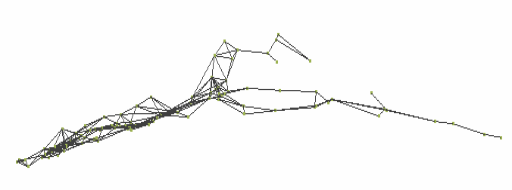}\put(0,30){C}\end{overpic}
\begin{overpic}[width=0.2\textwidth]{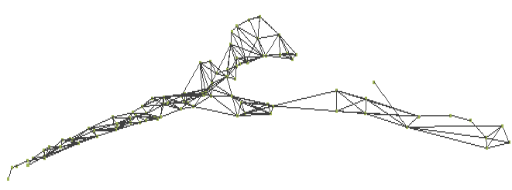}\put(0,30){D}\end{overpic}

\caption{The figure shows skeleton extraction results from different methods. A. Input 3D surface points; B. sampled skeleton points from surface points using DPC \cite{wu2015dpc}; C. skeleton mesh from surface points using Point2Skeleton \cite{lin2021point2skeleton}; D. skeleton mesh from our method with surface norm cost function. }
\label{fig:skeleton_results}
\end{figure*}

To evaluate our skeleton extraction method on Dataset 1, we carefully repair surface mesh using screened poisson surface reconstruction method \cite{kazhdan2013screened_poisson} with spherical harmonics to smooth the surface. Fig.\ref{fig:skeleton_results} shows the qualitative comparison of our methods and other state-of-the-art methods \cite{wu2015dpc,lin2021point2skeleton}.
Our method has better visual results. \cite{wu2015dpc} can generate unstructured skeleton points but it lacks topological constraint. We sample the number of points using \cite{poisson_sample} to be the same with the other methods for a fair comparison. It performs well when neuron has tube like structure but it is not good when neuron has a more circular shape. Compared with \cite{lin2021point2skeleton}, our method can capture more detailed structures which are important for sub-cellular feature extraction, such as branches. For a quantitative evaluation of our method on Dataset 1, we compute the strictly defined MAT and use the handcrafted method in \cite{lin2021point2skeleton} to remove insignificant spikes to get the simplified MAT. We sample points on the simplified MAT as the ground truth skeleton points.
Then we compute Chamfer Distance (CD) and Hausdorff distance (HD) between computed skeleton points and ground truth skeleton points. We refer to them as CD-skel and HD-skel, respectively. To compute CD-skel and HD-skel, we shift and rescale each skeleton so that the skeleton center is located at (0,0,0) and their x,y,z coordinates are all between -1 and 1 for every skeleton. We also measure the difference between the shapes reconstructed from the skeletons and ground truth surface points using CD and HD. We refer to them as CD-recon and HD-recon, respectively. Similarly, we also shift and rescale the ground truth points so that each neuron is centered at (0,0,0) and each neuron's surface coordinates are between -1 and 1. Other than those four aforementioned evaluation metrics, we also use the reconstructed neuron volume difference as the evaluation metric, considering neuron volume as one of the important property of a neuron. We denote it as $vol-pct$. Mathematically, it is defined as $vol-pct=\frac{|vrecon-vg|}{vg}$ where $vrecon$ is the volume of the reconstructed neurons from the skeleton model, and $vg$ is the ground truth volume. Table \ref{tab:D1_skeleton_extraction} gives the detailed results of different methods. It shows our method has the best performance compared to other methods on Dataset 1 in terms of all of the above evaluation metrics.

\begin{table*}[ht]
\scriptsize
    \centering
    \caption{Quantitative Comparison with state-of-the-art skeleton model extraction method on Dataset 1}
    \begin{tabular}{|c|c|c|c|c|c|}
    \hline
      str   &CD-recon& HD-recon&CD-skel&HD-skel&vol-pct ($\%$)\\
         \hline
         DPC \cite{wu2015dpc}& 0.102&0.298 &0.303&0.311&10.1\\
        \hline
        Point2Skeleton \cite{lin2021point2skeleton} & 0.081 &0.207 & 0.155&0.191&8.2 \\
        \hline
         Our method & \textbf{0.067} &\textbf{0.183} & \textbf{0.090}&\textbf{0.185}&\textbf{5.6}\\
         \hline
    \end{tabular}
    \label{tab:D1_skeleton_extraction}
\end{table*}

\subsection{Sub-cellular feature extraction from skeleton model}
We apply our sub-cellular feature extraction method on Dataset 1 and Dataset 2.
We define the length difference percentage (len-pct) and number of branches difference percentage (num-pct) to measure the neuron length error and number of branches error of the computation methods.
We compare our sub-cellular feature extraction method with the state-of-the-art sub-cellular feature extraction method proposed in \cite{skeleton_1_Navis}.
Table \ref{tab:sub-cellular feature results} gives details of sub-cellular feature computation results.
len-pct and num-pct for Dataset 1 is only for both animals. 
Our method provides the better sub-cellular feature extraction results in most cases and the percentage error is no more than 8 percent.
Also, as we see, our method is comparable to \cite{skeleton_1_Navis} on "curve" skeletons but has better performance on the skeleton meshes ("curve" skeleton is a special case of the skeleton mesh).
\begin{table*}[ht]
\scriptsize
    \centering
    \caption{Sub-cellular feature evaluation results}
    \begin{tabular}{|c|c|c|c|c|}
    \hline
         &len-pct on Dataset 1 ($\%$)& num-pct on Dataset 1 ($\%$)& len-pct on Dataset 2 ($\%$)&num-pct on Dataset 2 ($\%$)\\
         \hline
         NAVIS\cite{skeleton_1_Navis} & 7.1&10.5 &\textbf{3.2}&6.3\\
        \hline
         Our method & \textbf{5.5} &\textbf{7.6} & 3.8 &\textbf{5.8}\\
         \hline
    \end{tabular}
    \label{tab:sub-cellular feature results}
\end{table*}

\begin{figure*}[ht!]
    \centering
    \includegraphics[width=0.8\textwidth]{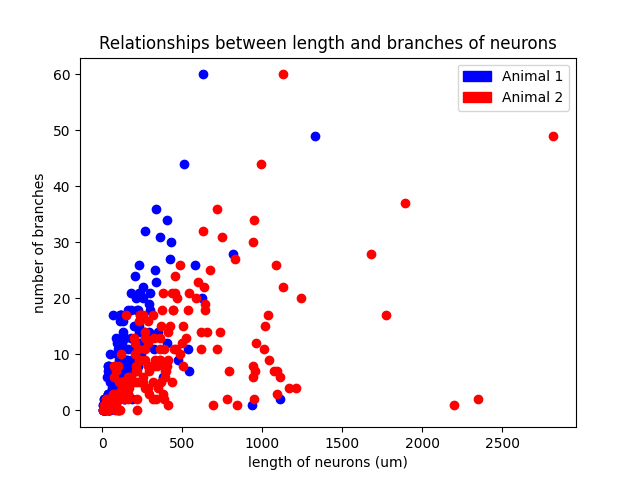}
    \caption{Relationships between length and number of branches of neurons using two animals of Dataset 1. Blue dots represent neurons from animal 1 and red dots represent neurons from animal 2.}
    \label{fig:len_branch}
\end{figure*}

For Dataset 1, we also analyze the relationships between length and branches of neurons computed from our method.
For animal 1 in Dataset 1, we first use our skeleton extraction method to get the skeleton representation of each neuron.
Next, we use our sub-cellular feature extraction method to get length of the neuron and number of branches of the neuron.
Fig. \ref{fig:len_branch} shows the relationships between length and branches of neurons. 
Blue dots represent neurons from animal 1 and red dots represent neurons from animal 2.
Overall, longer neurons tend to have more branches.
This relationship between neuron length and number of branches is consistent between animals 1 and 2.

\subsection{Skeleton Model Comparison}
We apply our skeleton model comparison method on Dataset 1 for the purpose of analyzing \textit{Ciona} neuron morphology of different neuron types. 
Given the skeleton graph, we embed it into a 100-dimension vector.
Then we use K-means++ to cluster vector representations of skeleton graphs.
For K-means++, the number of clusters is set to be the same as number of neuron classes.
After K-means++, the cluster label is given by using the majority vote of neuron types within the cluster.
We use animal 1 neurons to train the K-means++ model and get the cluster (neuron type) centers.
Next, we use animal 2 as the test set.
For each neuron in the test set, we assign the label based on its closest cluster center.
The distance metric we use is the euclidean distance in the vector embedding space.
Table \ref{tab:kmeans_clustering} shows the comparison of clustering (classification) accuracy on both training and test sets using different neuron classification method. The neuron classification methods include graph spectrum method, graph2vec \cite{narayanan2017graph2vec}, s-rep \cite{s_rep_object} and our graph level representation method.
The graph spectrum method uses the eigen values of the graph's adjacency matrix to form the vector representation.
Similar to our method, graph2vec method is another way to convert the skeleton graph to the graph level vector representation.
For the s-rep method, it uses the skeleton points' features such as spoke direction, spoke length, and skeleton points' locations to classify neurons.
From Table \ref{tab:kmeans_clustering}, we observe that grouping neurons by our graph embedding provide the best classification results on both train and test sets.
It shows that neuron types are closely related to its morphology.
Also, our method is a better way to represent skeleton graphs in terms of clustering accuracy.

\begin{table*}[ht]
\scriptsize
    \centering
    \caption{Neuron Classification Results}
    \begin{tabular}{|c|c|c|c|c|}
    \hline
         & Graph Spectrum& Graph2vec\cite{narayanan2017graph2vec} & S-rep\cite{s_rep_object}&Our method \\
         \hline
         Train& 0.691& 0.767&0.791&\textbf{0.893}\\
        \hline
         Test & 0.632 & 0.718 &0.773&\textbf{0.871}\\
         \hline
    \end{tabular}
    \label{tab:kmeans_clustering}
\end{table*}

\begin{figure*} [ht!]
\centering
\begin{overpic}[width=0.6\textwidth]{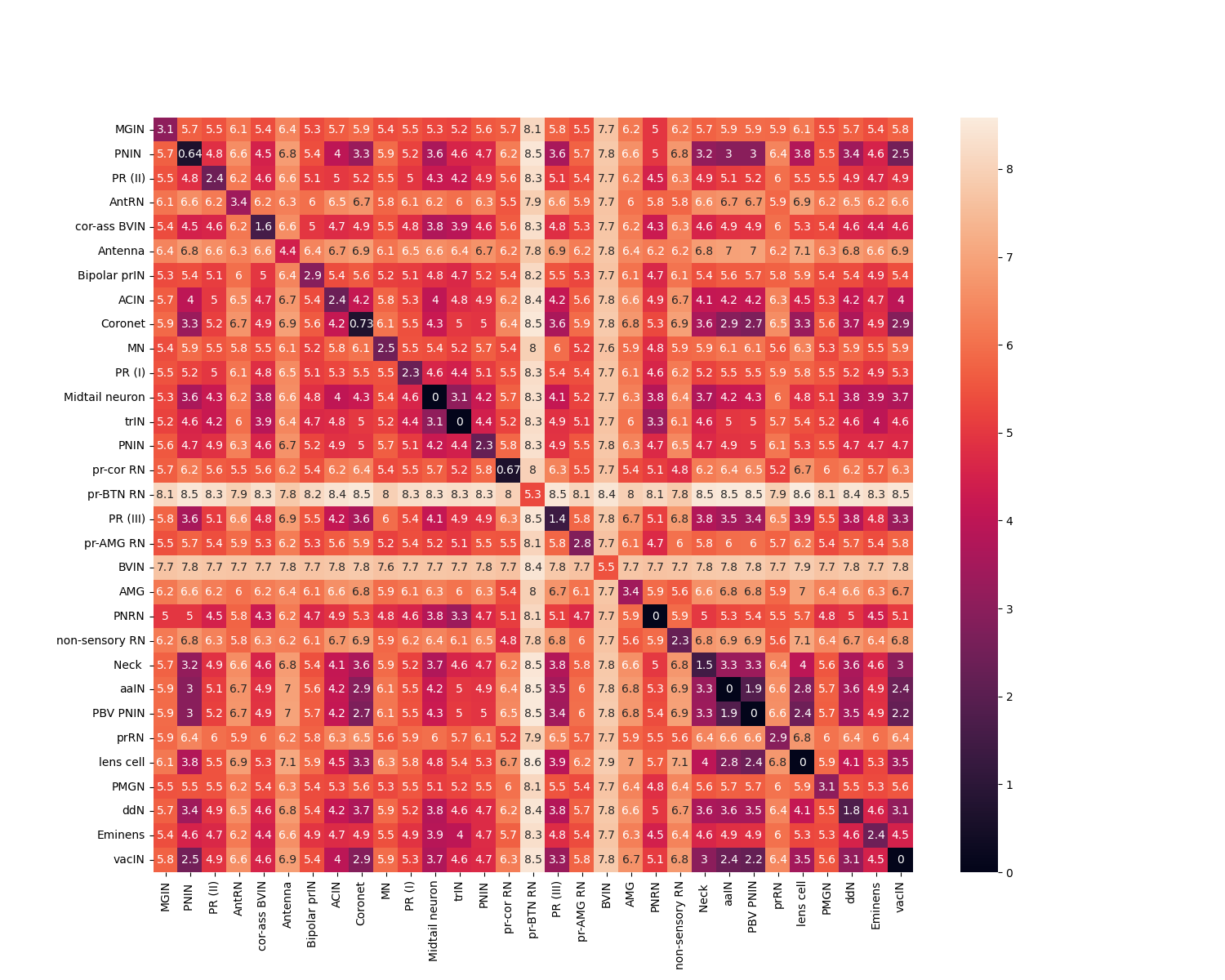}\put(5,4){A}\end{overpic}
\begin{overpic}[width=0.6\textwidth]{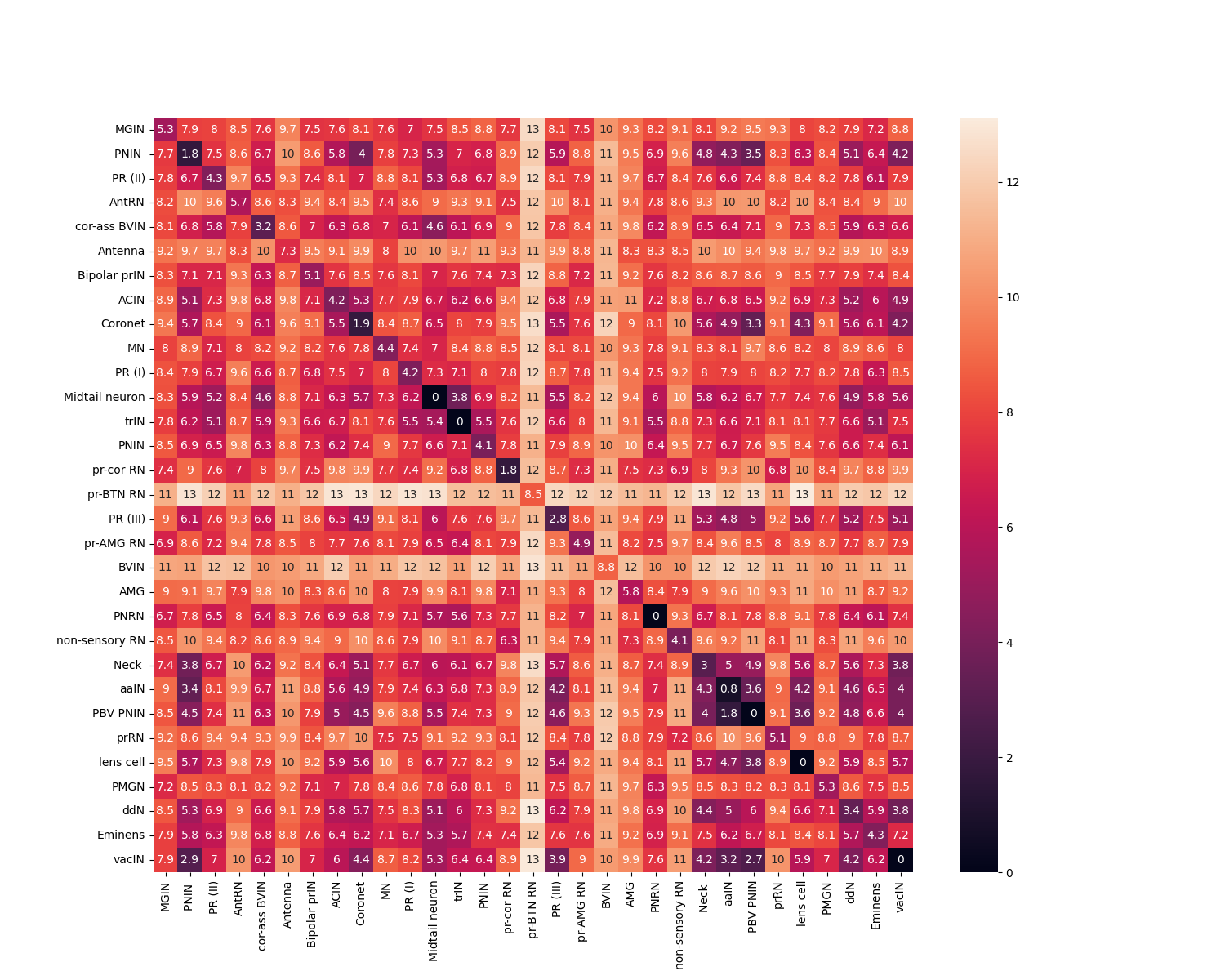}\put(5,4){B}\end{overpic}

\caption{Inter and intra class neuron morphology distance on animal 1 (A) and animal 2 (B) . Neuron morphology distance is computed by using euclidean distance between our graph level representation of the skeleton graph.}
\label{fig:pairwise_distance}
\end{figure*}

Based on previous observations, we do further morphology analysis based on our graph level representation results.
After we get the vector representation of each graph, we compute euclidean distance between each pair of vectors.
Then we compute the inter class and intra class distance based on pairwise neuron distance as Fig \ref{fig:pairwise_distance} shows.
Diagonal entries tend to be smaller than other values, confirming a strong correlation between structure and function.
More specifically, neurons within a neuron type tend to have a smaller morphological distance than neurons between different groups.
Also, two animals inter and intra distance look very similar.

\begin{figure*} [htp!]
\centering
\begin{overpic}[width=0.6\textwidth]{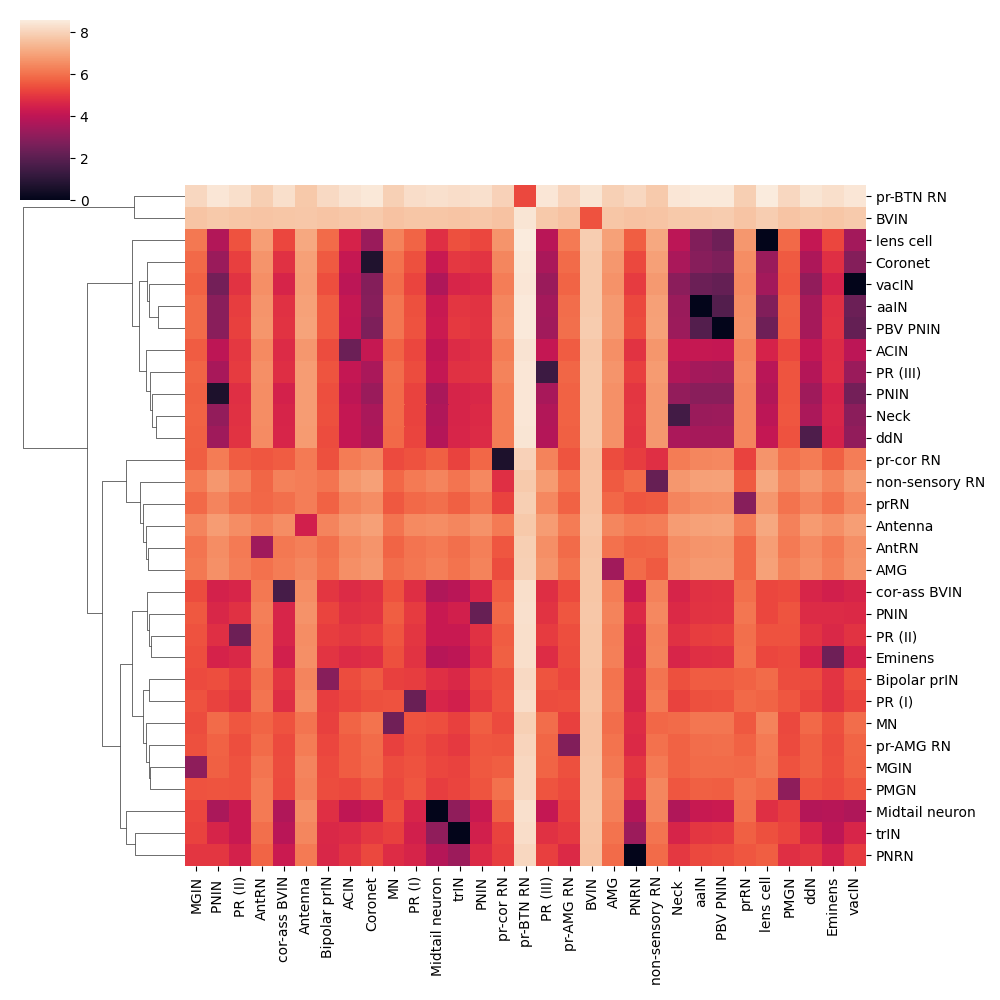}\put(5,4){A}\end{overpic}
\begin{overpic}[width=0.6\textwidth]{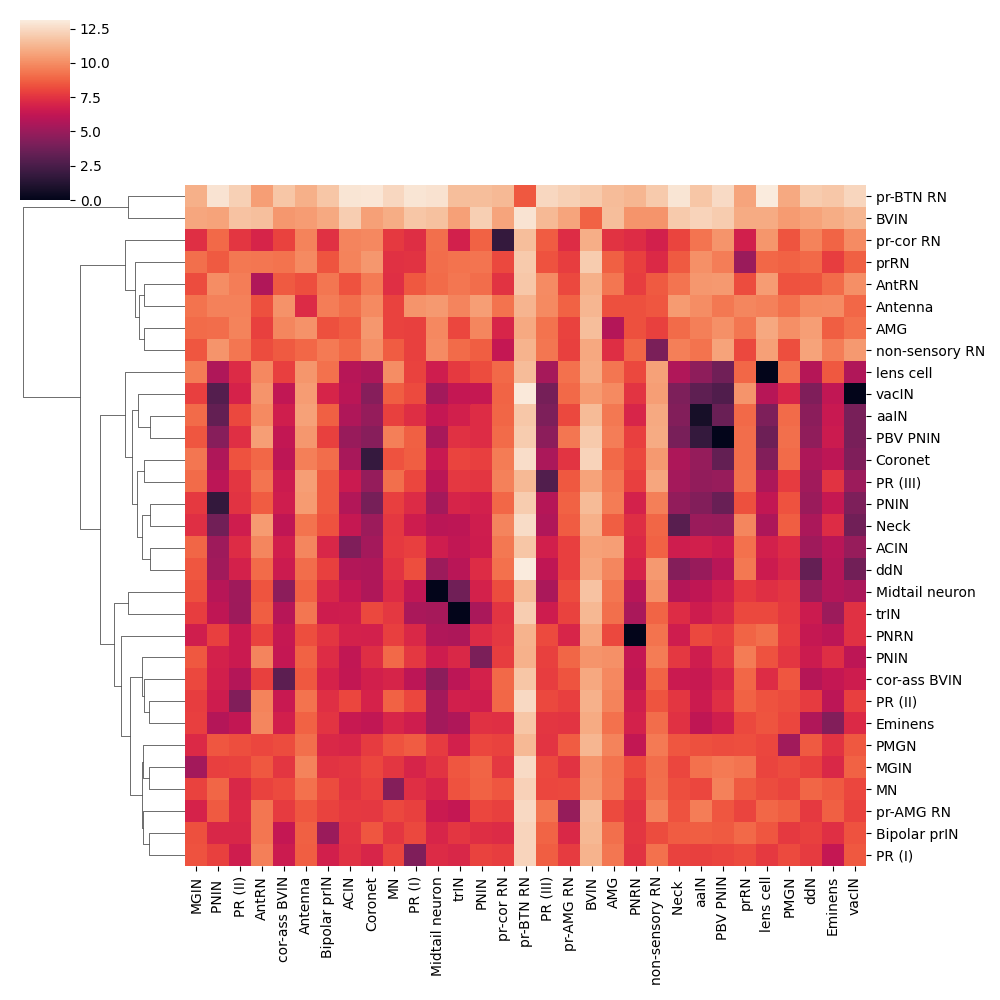}\put(5,4){B}\end{overpic}

\caption{Hierarchical clustering of neurons of animal 1 (A) and animal 2 (B).}
\label{fig:hierarchical_clustering}
\end{figure*}

Based on this inter and intra class distance, we do hierarchical clustering as shown in Fig \ref{fig:hierarchical_clustering}.
The hierarchical clustering results show that BVIN and pr-BTN RN have larger morphology distances from other neuron types.
The BVIN neurons are a broad group of intrinsic interneurons located in the brain vesicle of \textit{Ciona}.
The main role of this group is to connect the sensory neurons to other groups within the brain vesicle.
The BVIN neurons have partial subclassification based on sensory input \cite{kerrianne2016cns}. Receiving specific
sensory information is an indication of functional role, therefore, the BVIN can be further subdivided
into different groups based on the sensory group(s) from which they receive input. Using the sensory
input as criteria, the entire group was split up into four groups: prIN if receiving photoreceptor input,
antIN if receiving antenna cell input, pr-ant IN if receiving from both, and BVIN if not receiving from
either. The pr-BTN RN only have two neurons and their functions are mostly unknown. According to the
connectome \cite{kerrianne2016cns}, they receive input from both the photoreceptors and the BTN neurons (neurons
involved in processing touch stimuli in the tail), so it’s possible they play a role in integrating the two
inputs. Any functional differences that may exist between the two are currently unknown, however, the
hierarchical clustering suggests that this may be the case.

\section{Conclusion}
In this paper, we propose a novel neuron morphology analysis pipeline.
It mainly includes three parts.
First, we propose a robust shapre representation using skeleton mesh. Next, we compute sub-cellular features from the skeleton mesh. 
Finally, we compare different neuron shapes using skeleton mesh. To the best of knowledge, this is the first time that such an approach is used to represent and classify neuronal shapes.
 The introduction of the estimated surface norm penalty results in a robust mesh representation that achieves the state-of-the-art performance using well defined metrics. 
Based on skeleton graph, we formulate sub-cellular feature computation problem as a longest simple path problem that can be easily computed.
To compare different neuron morphology, we use a novel unsupervised graph level representation method to get the vector representation of each skeleton graphs.
We provide detailed experimental results to demonstrate the effectiveness of our method. 
Specifically, our analysis of the \textit{Ciona} dataset demonstrates that shape could be used as a significant feature for classifying neuronal types.

\section{Acknowledgement}
The authors would like to thank Kerrianne Ryan for providing the \textit{Ciona} EM dataset as well as the great annotations on the dataset. This work was supported
in part by grants from NIH \# 5R01NS103774-04 and NSF-SSI award \# 1664172.

\section{Author Contributions}
JJ designed the overall neuron morphology analysis pipeline, carried out experiments on different datasets, and drafted manuscript. MG helped on the clustering/classification part of the experiments and helped draft the manuscript.
CB provided the \textit{Ciona} neuron class information and helped draft the paper. WS coordinated \textit{Ciona} dataset acquisation, helped analyzed the \textit{Ciona} data, and reviewed the manuscript. BSM coordinated the overall design, development and evaluations of the neuron morphology analysis method, and helped prepare the manuscript.

\bibliographystyle{IEEEtran}
\bibliography{IEEEabrv,Bibliography}

\begin{thebibliography}{10}
\providecommand{\url}[1]{#1}
\csname url@rmstyle\endcsname
\providecommand{\newblock}{\relax}
\providecommand{\bibinfo}[2]{#2}
\providecommand\BIBentrySTDinterwordspacing{\spaceskip=0pt\relax}
\providecommand\BIBentryALTinterwordstretchfactor{4}
\providecommand\BIBentryALTinterwordspacing{\spaceskip=\fontdimen2\font plus
\BIBentryALTinterwordstretchfactor\fontdimen3\font minus
  \fontdimen4\font\relax}
\providecommand\BIBforeignlanguage[2]{{%
\expandafter\ifx\csname l@#1\endcsname\relax
\typeout{** WARNING: IEEEtran.bst: No hyphenation pattern has been}%
\typeout{** loaded for the language `#1'. Using the pattern for}%
\typeout{** the default language instead.}%
\else
\language=\csname l@#1\endcsname
\fi
#2}}

\bibitem{neuronmorphologyimpoartance}
M.~Halavi, K.~A. Hamilton, R.~Parekh, and G.~A. Ascoli, ``Digital
  reconstructions of neuronal morphology: three decades of research trends,''
  \emph{Frontiers in neuroscience}, vol.~6, p.~49, 2012.

\bibitem{neuron_tracing}
S.~Jiang, Z.~Pan, Z.~Feng, Y.~Guan, M.~Ren, Z.~Ding, S.~Chen, H.~Gong, Q.~Luo,
  and A.~Li, ``Skeleton optimization of neuronal morphology based on
  three-dimensional shape restrictions,'' \emph{BMC bioinformatics}, vol.~21,
  no.~1, pp. 1--16, 2020.

\bibitem{skeleton_1_Navis}
M.~Costa, J.~D. Manton, A.~D. Ostrovsky, S.~Prohaska, and G.~S. Jefferis,
  ``Nblast: rapid, sensitive comparison of neuronal structure and construction
  of neuron family databases,'' \emph{Neuron}, vol.~91, no.~2, pp. 293--311,
  2016.

\bibitem{skeleton_2_neuromopho}
M.~Abdellah, J.~Hernando, S.~Eilemann, S.~Lapere, N.~Antille, H.~Markram, and
  F.~Sch{\"u}rmann, ``Neuromorphovis: a collaborative framework for analysis
  and visualization of neuronal morphology skeletons reconstructed from
  microscopy stacks,'' \emph{Bioinformatics}, vol.~34, no.~13, pp. i574--i582,
  2018.

\bibitem{skeleton_3}
S.~Li, T.~Quan, C.~Xu, Q.~Huang, H.~Kang, Y.~Chen, A.~Li, L.~Fu, Q.~Luo,
  H.~Gong, \emph{et~al.}, ``Optimization of traced neuron skeleton using
  lasso-based model,'' \emph{Frontiers in neuroanatomy}, vol.~13, p.~18, 2019.

\bibitem{skeleton_method2}
P.~K. Saha, Y.~Xu, H.~Duan, A.~Heiner, and G.~Liang, ``Volumetric topological
  analysis: a novel approach for trabecular bone classification on the
  continuum between plates and rods,'' \emph{IEEE transactions on medical
  imaging}, vol.~29, no.~11, pp. 1821--1838, 2010.

\bibitem{unet_skeleton}
O.~Panichev and A.~Voloshyna, ``U-net based convolutional neural network for
  skeleton extraction,'' in \emph{Proceedings of the IEEE/CVF Conference on
  Computer Vision and Pattern Recognition Workshops}, 2019, pp. 0--0.

\bibitem{thinning_algorithm}
T.-C. Lee, R.~L. Kashyap, and C.-N. Chu, ``Building skeleton models via 3-d
  medial surface axis thinning algorithms,'' \emph{CVGIP: Graphical Models and
  Image Processing}, vol.~56, no.~6, pp. 462--478, 1994.

\bibitem{s_rep_object}
S.~M. Pizer, J.~Hong, J.~Vicory, Z.~Liu, J.~Marron, H.-y. Choi, J.~Damon,
  S.~Jung, B.~Paniagua, J.~Schulz, \emph{et~al.}, ``Object shape representation
  via skeletal models (s-reps) and statistical analysis,'' in \emph{Riemannian
  Geometric Statistics in Medical Image Analysis}.\hskip 1em plus 0.5em minus
  0.4em\relax Elsevier, 2020, pp. 233--271.

\bibitem{lin2021point2skeleton}
C.~Lin, C.~Li, Y.~Liu, N.~Chen, Y.-K. Choi, and W.~Wang, ``Point2skeleton:
  Learning skeletal representations from point clouds,'' in \emph{Proceedings
  of the IEEE/CVF Conference on Computer Vision and Pattern Recognition}, 2021,
  pp. 4277--4286.

\bibitem{sun2019infograph}
F.-Y. Sun, J.~Hoffman, V.~Verma, and J.~Tang, ``Infograph: Unsupervised and
  semi-supervised graph-level representation learning via mutual information
  maximization,'' in \emph{International Conference on Learning
  Representations}, 2019.

\bibitem{kerrianne2016cns}
K.~Ryan, Z.~Lu, and I.~A. Meinertzhagen, ``The cns connectome of a tadpole
  larva of ciona intestinalis (l.) highlights sidedness in the brain of a
  chordate sibling,'' \emph{Elife}, vol.~5, p. e16962, 2016.

\bibitem{scheffer2020connectome}
L.~K. Scheffer, C.~S. Xu, M.~Januszewski, Z.~Lu, S.-y. Takemura, K.~J.
  Hayworth, G.~B. Huang, K.~Shinomiya, J.~Maitlin-Shepard, S.~Berg,
  \emph{et~al.}, ``A connectome and analysis of the adult drosophila central
  brain,'' \emph{eLife}, vol.~9, p. e57443, 2020.

\bibitem{ascoli2007neuromorpho}
G.~A. Ascoli, D.~E. Donohue, and M.~Halavi, ``Neuromorpho. org: a central
  resource for neuronal morphologies,'' \emph{Journal of Neuroscience},
  vol.~27, no.~35, pp. 9247--9251, 2007.

\bibitem{qi2017pointnet++}
C.~R. Qi, L.~Yi, H.~Su, and L.~J. Guibas, ``Pointnet++: Deep hierarchical
  feature learning on point sets in a metric space,'' \emph{Advances in neural
  information processing systems}, vol.~30, 2017.

\bibitem{qi2017pointnet}
C.~R. Qi, H.~Su, K.~Mo, and L.~J. Guibas, ``Pointnet: Deep learning on point
  sets for 3d classification and segmentation,'' in \emph{Proceedings of the
  IEEE conference on computer vision and pattern recognition}, 2017, pp.
  652--660.

\bibitem{zhou2018open3d}
Q.-Y. Zhou, J.~Park, and V.~Koltun, ``Open3d: A modern library for 3d data
  processing,'' \emph{arXiv preprint arXiv:1801.09847}, 2018.

\bibitem{loss_GAE}
P.~V. Tran, ``Learning to make predictions on graphs with autoencoders,'' in
  \emph{2018 IEEE 5th international conference on data science and advanced
  analytics (DSAA)}.\hskip 1em plus 0.5em minus 0.4em\relax IEEE, 2018, pp.
  237--245.

\bibitem{cardona2012trakem2}
A.~Cardona, S.~Saalfeld, J.~Schindelin, I.~Arganda-Carreras, S.~Preibisch,
  M.~Longair, P.~Tomancak, V.~Hartenstein, and R.~J. Douglas, ``Trakem2
  software for neural circuit reconstruction,'' \emph{PloS one}, vol.~7, no.~6,
  p. e38011, 2012.

\bibitem{collins2007imagej}
T.~J. Collins, ``Imagej for microscopy,'' \emph{Biotechniques}, vol.~43,
  no.~S1, pp. S25--S30, 2007.

\bibitem{poisson_sample}
C.~Yuksel, ``Sample elimination for generating poisson disk sample sets,'' in
  \emph{Computer Graphics Forum}, vol.~34, no.~2.\hskip 1em plus 0.5em minus
  0.4em\relax Wiley Online Library, 2015, pp. 25--32.

\bibitem{wu2015dpc}
S.~Wu, H.~Huang, M.~Gong, M.~Zwicker, and D.~Cohen-Or, ``Deep points
  consolidation,'' \emph{ACM Transactions on Graphics (ToG)}, vol.~34, no.~6,
  pp. 1--13, 2015.

\bibitem{kazhdan2013screened_poisson}
M.~Kazhdan and H.~Hoppe, ``Screened poisson surface reconstruction,'' \emph{ACM
  Transactions on Graphics (ToG)}, vol.~32, no.~3, pp. 1--13, 2013.

\bibitem{narayanan2017graph2vec}
A.~Narayanan, M.~Chandramohan, R.~Venkatesan, L.~Chen, Y.~Liu, and S.~Jaiswal,
  ``graph2vec: Learning distributed representations of graphs,'' \emph{arXiv
  preprint arXiv:1707.05005}, 2017.

\end{thebibliography}

\vfill

\end{document}